\documentclass[10pt,conference]{IEEEtran}      
\IEEEoverridecommandlockouts
\usepackage{comment} %for comment
\usepackage{float}  % Add this to your preamble

\usepackage{afterpage} % Include this in the preamble
\usepackage{cite}
\usepackage{amsmath,amssymb,amsfonts}
\usepackage{algorithmic}
\usepackage{graphicx}
\usepackage{textcomp} 
\usepackage{mathtools}
\usepackage{csquotes}

\def\BibTeX{{\rm B\kern-.05em{\sc i\kern-.025em b}\kern-.08em
    T\kern-.1667em\lower.7ex\hbox{E}\kern-.125emX}}
\usepackage{graphicx, amsmath,mathrsfs, multirow, array, siunitx, rotating, booktabs, epstopdf, subfigure, amssymb, balance, hyperref, cite,soul}
\usepackage[lined, algoruled, commentsnumbered]{algorithm2e}
\usepackage[table]{xcolor}

\newcolumntype{L}[1]{>{\raggedright\let\newline\\\arraybackslash\hspace{0pt}}m{#1}}
\newcolumntype{C}[1]{>{\centering\let\newline\\\arraybackslash\hspace{0pt}}m{#1}}
\newcolumntype{R}[1]{>{\raggedleft\let\newline\\\arraybackslash\hspace{0pt}}m{#1}}
\usepackage[left=1.62cm,right=1.62cm,top=1.9cm]{geometry}
\definecolor{morange}{rgb}{0.8,0.2,0}
\definecolor{mblue}{rgb}{0,0.1,0.8}
\definecolor{mgreen}{rgb}{0,0.8,0.1}
\definecolor{mred}{rgb}{1,0,0}

\hypersetup{draft}

\usepackage{fancyhdr}   % Include the fancyhdr package
\usepackage{fancyhdr}
\usepackage{lipsum}

\newcommand{\IEEEcopyrightnotice}{%
  \footnotesize{© 2025 IEEE. Personal use of this material is permitted. 
  Permission from IEEE must be obtained for all other uses, in any current or future media, 
  including reprinting/republishing this material for advertising or promotional purposes, 
  creating new collective works, for resale or redistribution to servers or lists, 
  or reuse of any copyrighted component of this work in other works.}%
}

\fancypagestyle{firstpage}{%
  \fancyhf{}% Clear header/footer
  \fancyfoot[L]{\IEEEcopyrightnotice}% Center footer
}

\begin{document}

\title{Effective Feature Selection for Predicting Spreading Factor with ML in Large LoRaWAN-based Mobile IoT Networks}

\author{\IEEEauthorblockN{Aman Prakash\IEEEauthorrefmark{1}, Nikumani Choudhury\IEEEauthorrefmark{2}, Anakhi Hazarika\IEEEauthorrefmark{2}, Alekhya Gorrela\IEEEauthorrefmark{2}, } \IEEEauthorblockA{\IEEEauthorrefmark{1}
 National Institute of Advanced Manufacturing Technology, Ranchi, India } \IEEEauthorblockA{\IEEEauthorrefmark{2} 
Birla Institute of Technology \& Science, Pilani, Hyderabad, India}
Email: \{nikumani, anakhi.hazarika,  p20220103\}@hyderabad.bits-pilani.ac.in, amanprakash.connect@gmail.com}

\maketitle
\thispagestyle{firstpage}
\begin{abstract}
%IEEE24
%old abstract added result
\begin{comment}
LoRaWAN is a low-power long-range protocol that enables reliable and robust communication. This paper addresses the challenge of predicting the spreading factor (SF) in LoRaWAN networks using machine learning (ML) techniques. Optimal SF allocation is crucial for optimizing data transmission in IoT-enabled mobile devices, yet it remains a challenging task due to the fluctuation in environment and network conditions. We evaluated ML model performance across a large publicly available dataset to explore the best feature across key LoRaWAN features such as RSSI, SNR, frequency, distance between end devices and gateways, and antenna height of the end device.
To identify the top features, we trained and evaluated the model using k-nearest neighbors (k-NN), Decision Tree Classifier (DTC), Random Forest (RF), and Multinomial Logistic Regression (MLR) algorithms on 31 different combinations of 5 features mentioned above.
The finding of this paper provides valuable information for reducing the overall cost of dataset collection for ML model training by IoT devices and understanding the importance of specific features for optimizing SF prediction accuracy. This work contributes to a more reliable LoRaWAN system by offering a detailed analysis of feature importance.
    
\end{comment}
LoRaWAN is a low-power long-range protocol that enables reliable and robust communication. This paper addresses the challenge of predicting the spreading factor (SF) in LoRaWAN networks using machine learning (ML) techniques. Optimal SF allocation is crucial for optimizing data transmission in IoT-enabled mobile devices, yet it remains a challenging task due to the fluctuation in environment and network conditions. We evaluated ML model performance across a large publicly available dataset to explore the best feature across key LoRaWAN features such as RSSI, SNR, frequency, distance between end devices and gateways, and antenna height of the end device, further, we also experimented with 31 different combinations possible for 5 features. We trained and evaluated the model using k-nearest neighbors (k-NN), Decision Tree Classifier (DTC), Random Forest (RF), and Multinomial Logistic Regression (MLR) algorithms.
The combination of RSSI and SNR was identified as the best feature set. The finding of this paper provides valuable information for reducing the overall cost of dataset collection for ML model training and extending the battery life of LoRaWAN devices. This work contributes to a more reliable LoRaWAN system by understanding the importance of specific feature sets for optimized SF allocation.
%by offering a detailed analysis of feature importance.

\end{abstract}

\begin{IEEEkeywords}
LoRaWAN, Spreading Factor, Machine Learning, Random Forest, k-nearest neighbors, Multinomial Logistic Regression, Decision Tree Classifier
\end{IEEEkeywords}

\section{Introduction}

The Internet of Things (IoT) is a network of interconnected devices such as sensors, actuators, storage, and processing units with telecommunication interfaces. This allows the integration of any device with the internet, establishing interaction between devices that are commonly referred to as machine-to-machine (M2M) communications \cite{1STBOOK}.
%The number of IoT devices has grown exponentially in the last couple of years \cite{Intro_1ST_LINE}.
%Therefore,  
Low Power Wide Area Network (LPWANs) is commonly used for static and mobile IoT devices that provide multi-year battery life and send small amounts of data over long distances with the transmission frequency being a few times per hour with varying environments \cite{LoRaWAN_official_website}. LoRa is one of the prime communication standards requiring low power for long-range communications for IoT devices. However, this network faces scalability, low data rate, and other performance issues.

Data rate (DR), Spreading Factor (SF), and bandwidth (BW) are the parameters responsible for efficient data transmission in LoRaWAN. 
The SF is a tunable parameter that controls the speed of data transmission, where, lower SF indicates higher DR. Allocating optimal SF overcomes these issues and helps to improve the performance of the network \cite{Intro_SF, isfa}.

\begin{figure}
    \centering
    \includegraphics[width=0.9\linewidth]{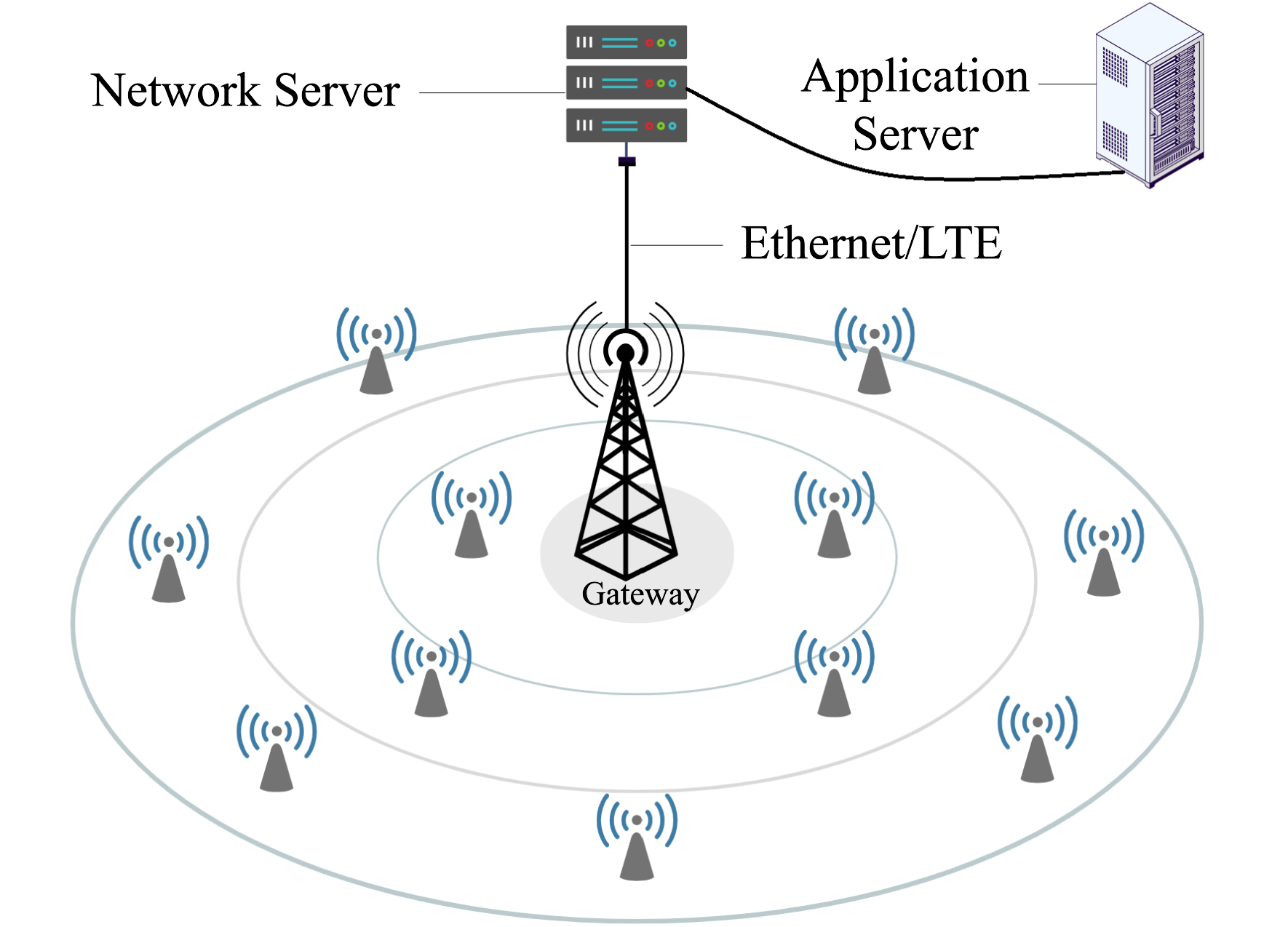}
    \caption{LoRaWAN network architecture}
    \label{fig:a}
\end{figure}

\subsection{\textit{LoRaWAN}}
\label{LoRa}
LoRa developed by Semtech Corporation is a Radio Frequency modulation technology for low-power, long-range networks. Its communication range is up to three miles in urban areas and up to ten miles in rural areas. A key characteristic of LoRa is its very low power consumption which enables battery-powered devices to operate for up to ten years \cite{semtech_offcial_Website}. %IEEE24    REMOVED: The signal emitted by the LoRa modulation is a symbol called Chirp. Its name "Chirp" comes from the fact that this symbol is used in radar technology (Chirp: Compressed High Intensity Radar Pulse).
LoRaWAN is an extension of LoRa, developed by LoRa Alliance. LoRaWAN has several benefits over LoRa, such as end-to-end encryption, bi-directional communication, and open standards. LoRa Alliance is a non-profit organization focused on maintaining and encouraging LoRaWAN's open ecosystem globally \cite{1STBOOK}.

LoRaWAN architecture comprises a Gateway (GW), end devices (EDs), network server, and application server\cite{LoRaWAN_official_website} as illustrated in Figure \ref{fig:a}. Various non-ML methods are available in the literature for optimal SF selection. For instance, for static EDs (e.g., water meter, gas monitoring), Adaptive Data Rate (ADR) protocol is used, and for mobile EDs (e.g., pet tracking), algorithms like Blind ADR \cite{Blind_ADR_Downloadable}, Enhanced ADR \cite{EADR}, and MADERE \cite{madere} are used. These methods for SF selection often rely on heuristic approaches and are insufficient to adapt to dynamic large IoT scenarios.

\subsection{\textit{Spreading Factor (SF)}}

The number of bits encoded in a symbol by LoRa is an adaptable resource parameter called SF. There are a total of 6 SFs i.e., (SF7-SF12) operated by LoRa. The greater the SF (e.g. SF11, SF12), the farther the network distance coverage, lower battery power usage and data rate, the inverse is true for lower SFs. Hence, SF choice plays a significant role in overall network performance.
\cite{semtech_offcial_Website}.

With a remarkable amount of data and computing resources available, there is widespread interest in applying ML methods to communication system solutions \cite{IEEE_invited_paper_ml}. In machine learning, a feature is an individual measurable property that serves as an input variable for training of machine learning models, playing a critical role in the model's ability to make predictions. Despite extensive studies on SF prediction using ML in LoRaWAN networks, a research gap exists in identifying the optimal combination of input features to improve SF prediction accuracy. The major contributions of this work are as follows:
\begin{enumerate}
    \item Develop a feature selection method to identify the best set of LoRaWAN features for predicting optimal SF allocation.
    \item Design and train different ML algorithms for 31 unique combinations of features present in publicly available LoRaWAN dataset.
    \item Accuracy and F1 Score metrics are analyzed to evaluate the performance of ML models on different combinations of input features.  
\end{enumerate}

% In this research, we aim to enhance the predictability of SF by identifying the most useful set of input features used to train an ML model. We did this evaluation by training classification ML models for different combinations of five features present in the large, publicly available LoRaWAN dataset.

The rest of the section of the paper is as follows. Section~\ref{related} details related work in Machine Learning. Section~\ref{methods} describes the dataset, feature selection and combination approach, ML algorithms, evaluation metrics, and computing specifications, followed by Section~\ref{results}, which presents the key findings and their implications. We conclude in Section~\ref{conclude} with a discussion on future work.
\vspace{\baselineskip}
%\section{LoRa/LoRaWAN}

\vspace{\baselineskip}
\section{Related Work}
\label{related}

The authors in \cite{13feature-oneHotEncoding} used various ML algorithms to predict RSSI using 13 features. Algorithms such as linear regression, polynomial regression of degree 3 (expanding 13 features into 560 to capture non-linear relationships), k-NN, Random Forest, and Gradient Boost. The Gradient Boost model achieved the lowest prediction error leading to the conclusion that latitude, longitude and altitude are the most significant features in predicting RSSI.

%REMOVED
%The authors in~\cite{commonfeature5} explored ML methods to assign optimal SF. They used a simulated dataset containing three features (X and Y coordinates of transmission source and SF of transmission) to train models using an SVM and DTC algorithm. They demonstrated the enhancement of network performance by managing network collisions of EDs. 

This study \cite{L1-related-study} used various ML techniques for signal loss prediction in LoRa. Lasso, one of the explored techniques, automatically selects the most relevant features and discards others, leading to a simplified model, it achieved RMSE(dB) of 9.41 outperforming several classical algorithms.

%inspirational paper 
The authors in \cite{relatedwork_aus} experimented with four sets of training features (i.e., RSSI, RSSI + SNR, RSSI + SF, RSSI + SF + SNR) to train ML models using five different algorithms %(i.e., Random Forest (RF), k-nearest neighbors (k-NN), Support Vector Regression (SVR), Gradient Boost (GB) and Multi-layer perception (MLP))
to estimate the distance of a target node from a LoRa gateway. They used their privately collected data and a public dataset to evaluate the results using the mean average error metric. RSSI + SF + SNR training feature outperforms other features. 

The authors in \cite{commonfeature4} propose a deep learning-based novel approach-"AI-ERA" to assign SF to both static and mobile devices proactively. They generated the dataset using ns-3 to train a deep neural network. This approach improved the packet success ratio by 32\% and 28\% for static and mobile EDs respectively.

Many studies focused on novel approaches for SF prediction using ML, but there remains a critical gap in understanding which input features are most important for model training. To the best of our knowledge, this is the first comprehensive ranking feature importance and identifying the optimal feature set focused on SF in LoRaWAN networks.

\vspace{\baselineskip}
\section{Methodology}
\label{methods}

\subsection{LoRaWAN Dataset}

\subsubsection{Dataset Description}

%IEEE24
%added sth, tells we used others dataset in study
This study utilizes a comprehensive LoRaWAN dataset made available through \cite{DATASET_PAPER}, consisting of 930,753 datapoints and 25 features in a comma-separated value (.csv) file recorded from October 2021 to March 2022 with an average sampling duration of 60 seconds. All features of the dataset including SF are listed in Table \ref{tab:compact_parameters}.

\begin{table}[h!]
\centering
\caption{Summary of Dataset features}
\label{tab:compact_parameters}
\begin{tabular}{|l|l|}
\hline

Index & Carrier Frequency \\ \hline
Timestamp & Frame Length \\ \hline
Antenna height (of GW) & Temperature \\ \hline
Antenna height (of ED) & Relative Humidity \\ \hline
ED ID & Pressure \\ \hline
 Experimental path loss  & PM2.5 \\ \hline
Distance between the GW and ED & RSSI \\ \hline
ED Transmitter Radiated Power & SNR \\ \hline
ED Transmitter Losses & Time on Air \\ \hline
ED Transmitter Antenna gain & \textbf{SF} \\ \hline
GW Losses & Noise Power  \\ \hline
GW Antenna gain & Signal Power \\ \hline
 Energy consumed by the transmission & \\ \hline
\end{tabular}

\end{table}

\subsubsection{Data Collection Setup}
The experimental setup for the collection of the dataset consists of four EDs and one GW located in the urban area of Medellín, Colombia. The EDs were strategically placed at varying distances from GW in the city. The GW was connected to the internet using Ethernet. All the collected data was stored in a cloud MySQL database.
\vspace{\baselineskip}

\subsection{Feature Selection and Combination Approach}

One of the preliminary steps in  ML model training is input feature selection.  In ML literature, there exist several traditional statistical methods to rank features based on their ability to enhance the predictive accuracy of the target variable, such as filter methods, wrapper methods, and embedded methods \cite{PCC-Filter}. However, to interpret the results while applying domain knowledge of LoRaWAN, we adopt our straightforward approach.

\subsubsection{Picking Features}
From the LoRaWAN dataset, we selected five key features out of 24 and one target variable i.e. SF. 
%IEEE24; signifies that below 4 studies cited talks about SF prediction not feature selection
These features were chosen based on their theoretical relevance to LoRaWAN performance and their common usage in existing studies exploring different approaches for SF prediction 
\cite{commonfeature1}\cite{commonfeature2}\cite{commonfeature4}\cite{commonfeature5}:
\begin{itemize}
    \item Distance between the GW and EDs
    \item Antenna height (ED)
    \item Carrier frequency of the transmission
    \item RSSI (Received Signal Strength Indicator)
    \item SNR (Signal-to-Noise Ratio)

\end{itemize}

Selecting these five features allows us to constrain the feature set. 
%IEEE24; removed below line
%Future research can expand this study by including more features.
%give 1 line space
\vspace{\baselineskip}

\subsubsection{Feature Combination Approach}
For further exploration of the impact of features on model performance, we incorporated a feature combination approach. We generated all possible combinations of five features. For illustration, examples of some feature combinations include:
\begin{itemize}
    \item Combination 1: RSSI
    \item Combination 2: RSSI + SNR
    \item Combination 3: Frequency + SNR +  RSSI
\end{itemize}
%originalBeforeIEEE
%IEEE24; removed childish calculations
\begin{comment}
    The total number of feature combinations is calculated as follows:
\[
\binom{5}{1} + \binom{5}{2} + \binom{5}{3} + \binom{5}{4} + \binom{5}{5}
\]
Adding these, we get:
\[5+10+10+5+1=31\]
Alternatively, we can use the combinatorics formula: 
\[2^n - 1\]
where n is the number of features. For n=5, we get
\[2^{5} - 1 = 31\]
Each of these 31 combinations of features is used to train the model with four different ML algorithms, resulting in this number of models:
\[31\times4=124\]
\end{comment}

\noindent The total number of feature combinations is given by the combinatorics formula:
\begin{equation}
\sum_{k=1}^{n} \binom{n}{k} = 2^n - 1\
\end{equation}
where \( n \) is the number of features. For \( n=5 \), we get 31.

\noindent Each of these 31 combinations is used to train the model with four different ML algorithms, resulting in 124 models.

\noindent Details of all feature combinations and their performances are given in section ~\ref{results}.

\vspace{\baselineskip}

\subsection{Machine Learning Models}
%\vspace{\baselineskip}

\noindent We utilized four different classification ML algorithms, that is: k-Nearest-Neighbours (k-NN), Decision Tree Classifier, Multinomial Logistic Regression, and Random Forest to evaluate the performance of the selected feature. These algorithms were found to be the most frequently applied in existing studies for improving the performance of LoRa and LoRaWAN \cite{commonfeature2}\cite{commonfeature5} \cite{refrence3} \cite{mayberelatedstudy}.
Labelled data was divided into training set (80\%) and rest (20\%) for testing of ML models. The results for each model were recorded to identify the most important feature combination and algorithm for predicting the Spreading Factor (SF). The best k value for the k-NN algorithm was chosen based on the highest weighted F1 score. Table \ref{tab:hyperparameters} contains the specific hyperparameters for different algorithms used in this research.

\begin{table}[h]
\centering
\caption{Hyperparameters for four ML Algorithms}
\label{tab:hyperparameters}
\begin{tabular}{|l|l|}
\hline
\textbf{Algorithm} & \textbf{Hyperparameter} \\ \hline
k-nearest neighbors (k-NN) & k (Number of neighbors): 1 to 20 \\ \hline
Decision Tree Classifier & Random State: 42 \\ \hline
Multinomial Logistic Regression & Solver: lbfgs \\ 
 & Max Iterations: 1000 \\ 
 & Random State: 42 \\ \hline
Random Forest & Estimators Number: 100 \\ 
 & Bootstrap: True \\ 
 & Random State: 42 \\ \hline
\end{tabular}
\abovecaptionskip=10pt

\end{table}

\subsection{Evaluation Metrics}

 In this study, we used accuracy and F1 Score metrics to evaluate the performance of machine learning models. Accuracy alone can be sometimes misleading for an imbalanced dataset hence another metric F1 score also taken into consideration which can help to better conclude the results. The F1 Score is the harmonic mean of precision and recall \cite{eval_metric_defination2}. Precision and recall are just two of many classification measures. Both the metrics provide a different perspective on the model prediction and help us understand the overall performance. 
 
 \noindent These metrics are computed for each feature combination of all four machine learning algorithms used in this study.

%IEEE24
%commented accuracy definition
\begin{comment}
\subsubsection{Accuracy}
It generally measures how well the model is performing. 

It is defined as:
\begin{equation}
\text{Accuracy} = \frac{TP + TN}{TP + TN + FP + FN}
\end{equation}
TN, TP, FN, and FP stand for \textit{True Negative, True Positive, False Negative, and False Positive} respectively.

\vspace{\baselineskip}
\subsubsection{F1 Score}
This single metric provides an overview of two other metrics namely, precision and recall. Precision and recall are just two of many classification measures derived from TP, FP, TN, and FN \cite{eval_metric_defination1}. Precision and recall are defined as:
 
\begin{equation}
\text{Precision} = \frac{TP}{TP + FP}
\end{equation}

\begin{equation}
\text{Recall} = \frac{TP}{TP + FN}
\end{equation}

%leaves no indent
\noindent The F1 Score is the harmonic mean of precision and recall. Given as: 

\begin{equation}
F1 = \frac{2 \cdot \text{Precision} \cdot \text{Recall}}{\text{Precision} + \text{Recall}}
\end{equation}
\end{comment}
\vspace{\baselineskip}
 \subsection{Computing Specifications}
 \noindent The training of machine learning models was performed on a desktop computer, with an 11th Gen Intel(R) Core(TM) i5-11300H processor and 16.0 GB installed RAM capacity. 
 
\noindent The ML models were trained using the following software and package versions:
\begin{itemize}
    \item Python: 3.11.4 (packaged by Anaconda, Inc.)
    \item Jupyter Notebook: 6.5.7
    \item Anaconda: 23.7.2
    \item scikit-learn: 1.3.0
\end{itemize}
\vspace{\baselineskip}
\vspace{\baselineskip}

\section{Experimental Results and Analysis}
\label{results}

%IEEE24
%Modified Little bit below para
The analysis of the accuracy and F1 Score of the trained ML models provided several key insights and trends for the prediction of SF. This analysis includes the best and worst-performing ML algorithms, the impact of different features and the general trend of accuracy and F1 score as the number of features in combinations increases. To validate our feature analysis findings, we also employed Pearson's correlation. All the metrics calculated are summarised in Table~\ref{tab:single-features}-~\ref{tab:five-features}. The code employed in this research is available on GitHub \cite{Github-code}.

\subsection{Feature Analysis}
\label{feature analysis}
The analysis reveals that the combinations, including the RSSI feature, outperform other features. Single RSSI features have higher accuracy and F1 score than other single features as observed from Table~\ref{tab:single-features}. A similar trend is noticed when RSSI is combined with other features across all algorithms, whether in two, three, or four-feature combinations. This corroborates with the working principle of the ADR technique in LoRaWAN networks. 
SNR was found to be the second most important feature after RSSI.  Together RSSI and SNR combination performed better than either feature alone, also it outperformed every other two feature combinations, as evident from Table~\ref{tab:two-features}. The antenna height of end devices and distance between the end devices and gateway feature showed moderate performance as individual performance. However, when combined with RSSI then its performance slightly increased.

The frequency feature was identified as the lowest performer among all individual features. Models that included frequency in combination with other features had lower performance compared to those that included RSSI or SNR. 
This may be because, in LoRaWAN networks, the carrier frequency is a fixed parameter that varies primarily by region. For instance, in Europe, the frequency bandwidth is set between 863 MHz to 870 MHz~\cite{1STBOOK}. Given its limited variation within a specific deployment, the frequency may be leading to lower performance. However, when the frequency is combined with both  RSSI and SNR, a noticeable improvement is observed, as seen in Table~\ref{tab:three-features}.
%redundant
%From Tables~\ref{tab:three-features} and ~\ref{tab:four-features}, we can notice all the feature combinations having both RSSI and SNR achieve an average accuracy and F1 scores close to 65\%, while the rest fall below 60\%.

%IEEE24; added below para
%Another Strong result in the below para
 The average percentages of accuracy and F1-score, calculated across four machine learning algorithms for all 31 feature combinations are visualized in Figure 2, with feature combination serial numbers (detailed in Table~\ref{tab:single-features}-~\ref{tab:five-features}) on the x-axis and corresponding metrics on the y-axis. Its analysis reveals an upward trend in performance metrics as the number of features in combinations increases. Notably, feature combinations 6 (RSSI + SNR), 16 (RSSI + SNR + Distance), 17 (RSSI + SNR + Height), 18 (RSSI + SNR + Frequency), 26 (RSSI + SNR + Distance + Height), 27 (RSSI + SNR + Distance + Frequency), 28 (RSSI + SNR + Frequency + Height), and 31 (all five features) consistently achieve the highest performance metrics, approaching 65\% while the rest fall below 60\% as visualized in Figure 2, a common observation is that all of these feature combinations include RSSI and SNR as features which validate these two as top features. The analysis indicates that the difference in average metrics when using RSSI + SNR alone (feature combination 6) is minimal compared to using RSSI + SNR alongside additional features ( three, four, or five features). Notably, this is the smallest feature combination set among the above best-performing options. This suggests that if the ML model is trained on (RSSI + SNR) set, there would be a negligible impact on the quality of SF prediction over the models utilizing all five features, which have the highest average metric, approaching 65.5 \%.

\begin{table*}[htbp]
\centering
\caption{Performance Metrics for Single Features}
\label{tab:single-features}
\begin{tabular}{|c|l|cc|cc|cc|cc|cc|}
\hline
\multirow{2}{*}{No.} & \multirow{2}{*}{Feature} & \multicolumn{2}{c|}{k-NN} & \multicolumn{2}{c|}{MLR} & \multicolumn{2}{c|}{DTC} & \multicolumn{2}{c|}{RF} & \multicolumn{2}{c|}{Average} \\
 &  & Acc \% & F1 & Acc \% & F1 & Acc \% & F1 & Acc \% & F1 & Acc \% & F1 \\
\hline
1 & RSSI & 57.81 & 60.65 & 52.04 & 43.02 & 59.44 & 58.35 & 59.44 & 58.35 & 57.18 & 55.07 \\
2 & SNR & 54.36 & 57.52 & 52.83 & 45.69 & 56.94 & 52.14 & 56.94 & 52.14 & 55.27 & 51.87 \\
3 & Frequency & 44.60 & 37.65 & 52.53 & 36.19 & 52.53 & 36.19 & 52.53 & 36.19 & 50.55 & 36.55 \\
4 & Height & 52.54 & 56.77 & 52.60 & 42.47 & 52.94 & 49.23 & 52.94 & 49.23 & 52.76 & 49.43 \\
5 & Distance & 53.12 & 53.96 & 52.34 & 36.19 & 52.94 & 49.23 & 52.94 & 49.23 & 52.88 & 47.15 \\
\hline
\end{tabular}
\end{table*}
 
\begin{table*}[htbp]
\centering
\caption{Performance Metrics for Two-Feature Combinations}
\label{tab:two-features}
\begin{tabular}{|c|l|cc|cc|cc|cc|cc|}
\hline
\multirow{2}{*}{No.} & \multirow{2}{*}{Features} & \multicolumn{2}{c|}{k-NN} & \multicolumn{2}{c|}{MLR} & \multicolumn{2}{c|}{DTC} & \multicolumn{2}{c|}{RF} & \multicolumn{2}{c|}{Average} \\
 &  & Acc \% & F1 & Acc \% & F1 & Acc \% & F1 & Acc \% & F1 & Acc \% & F1 \\
\hline
6 &\textbf{RSSI+SNR} & \textbf{64.43} & \textbf{65.57} & \textbf{59.69} & \textbf{57.02} & \textbf{66.23} & \textbf{66.71} & \textbf{66.21} & \textbf{66.58} & \textbf{64.14} & \textbf{63.97} \\
7 & RSSI+Frequency & 58.33 & 60.63 & 52.03 & 43.04 & 60.05 & 60.41 & 60.07 & 60.55 & 57.62 & 56.16 \\
8 & RSSI+Height & 59.59 & 61.76 & 54.66 & 47.58 & 61.31 & 61.02 & 61.31 & 61.02 & 59.22 & 57.84 \\
9 & RSSI+Distance & 59.83 & 62.00 & 52.73 & 45.76 & 61.31 & 61.02 & 61.31 & 61.02 & 58.79 & 57.45 \\
10 & SNR+Frequency & 54.38 & 57.03 & 52.86 & 45.89 & 57.63 & 55.57 & 57.32 & 55.80 & 55.63 & 53.57 \\
11 & SNR+Height & 55.61 & 53.12 & 53.31 & 48.34 & 58.16 & 55.65 & 58.17 & 55.33 & 56.31 & 53.11 \\
12 & SNR+Distance & 56.03 & 57.87 & 53.25 & 49.05 & 58.16 & 55.65 & 58.17 & 55.33 & 56.40 & 54.48 \\
13 & Frequency+Height & 52.87 & 56.51 & 52.66 & 44.06 & 53.10 & 53.64 & 53.10 & 53.64 & 52.93 & 51.96 \\
14 & Frequency+Distance & 52.87 & 56.51 & 52.53 & 36.19 & 53.10 & 53.64 & 53.10 & 53.64 & 52.90 & 50.00 \\
15 & Height+Distance & 53.12 & 53.96 & 52.60 & 42.47 & 52.94 & 49.23 & 52.94 & 49.23 & 52.90 & 48.72 \\
\hline
\end{tabular}
\end{table*}

\begin{table*}[htbp]
\centering
\caption{Performance Metrics for Three-Feature Combinations}
\label{tab:three-features}
\begin{tabular}{|c|l|cc|cc|cc|cc|cc|}
\hline
\multirow{2}{*}{No.} & \multirow{2}{*}{Features} & \multicolumn{2}{c|}{k-NN} & \multicolumn{2}{c|}{MLR} & \multicolumn{2}{c|}{DTC} & \multicolumn{2}{c|}{RF} & \multicolumn{2}{c|}{Average} \\
 &  & Acc \% & F1 & Acc \% & F1 & Acc \% & F1 & Acc \% & F1 & Acc \% & F1 \\
\hline
16 & RSSI+SNR+Distance & 65.14 & 65.87 & 60.05 & 58.89 & 66.60 & 67.14 & 66.59 & 67.04 & 64.59 & 64.73 \\
17 & RSSI+SNR+Height & 65.08 & 65.78 & 59.51 & 57.17 & 66.60 & 67.14 & 66.59 & 67.04 & 64.45 & 64.28 \\
18 & RSSI+SNR+Frequency & 66.25 & 66.52 & 59.53 & 56.78 & 67.73 & 67.76 & 67.75 & 67.75 & 65.32 & 64.70 \\
19 & RSSI+Distance+Height & 59.83 & 62.00 & 54.50 & 48.28 & 61.31 & 61.02 & 61.31 & 61.02 & 59.24 & 58.08 \\
20 & RSSI+Distance+Frequency & 59.75 & 60.90 & 52.38 & 44.68 & 61.77 & 61.83 & 61.77 & 61.71 & 58.92 & 57.28 \\
21 & RSSI+Height+Frequency & 59.59 & 60.74 & 54.78 & 47.71 & 61.77 & 61.83 & 61.77 & 61.71 & 59.48 & 58.00 \\
22 & SNR+Distance+Height & 56.03 & 57.87 & 53.55 & 49.53 & 58.16 & 55.65 & 58.17 & 55.33 & 56.48 & 54.60 \\
23 & SNR+Distance+Frequency & 56.71 & 58.03 & 53.27 & 49.53 & 58.76 & 57.17 & 58.76 & 57.17 & 56.88 & 55.47 \\
24 & SNR+Frequency+Height & 55.39 & 58.04 & 53.38 & 48.62 & 58.76 & 57.17 & 58.76 & 57.17 & 56.57 & 55.25 \\
25 & Frequency+Distance+Height & 52.87 & 56.51 & 52.66 & 44.06 & 53.10 & 53.64 & 53.10 & 53.64 & 52.93 & 51.96 \\
\hline
\end{tabular}
\end{table*}

\begin{table*}[htbp]
\centering
\caption{Performance Metrics for Four-Feature Combinations}
\label{tab:four-features}
\begin{tabular}{|c|l|cc|cc|cc|cc|cc|}
\hline
\multirow{2}{*}{No.} & \multirow{2}{*}{Features} & \multicolumn{2}{c|}{k-NN} & \multicolumn{2}{c|}{MLR} & \multicolumn{2}{c|}{DTC} & \multicolumn{2}{c|}{RF} & \multicolumn{2}{c|}{Average} \\
 &  & Acc \% & F1 & Acc \% & F1 & Acc \% & F1 & Acc \% & F1 & Acc \% & F1 \\
\hline
26 & RSSI+SNR+Distance+Height & 65.14 & 65.87 & 60.34 & 59.24 & 66.60 & 67.14 & 66.59 & 67.03 & 64.67 & 64.82 \\
27 & RSSI+SNR+Distance+Frequency & 66.48 & 66.76 & 60.11 & 59.00 & 68.04 & 68.05 & 68.05 & 67.99 & 65.67 & 65.45 \\
28 & RSSI+SNR+Frequency+Height & 66.71 & 66.95 & 59.38 & 57.06 & 68.04 & 68.05 & 68.05 & 67.99 & 65.55 & 65.01 \\
29 & RSSI+Frequency+Distance+Height & 59.75 & 60.90 & 54.23 & 47.76 & 61.77 & 61.83 & 61.77 & 61.71 & 59.38 & 58.05 \\
30 & Frequency+SNR+Distance+Height & 56.71 & 58.03 & 53.56 & 49.74 & 58.78 & 57.17 & 58.76 & 57.17 & 56.95 & 55.53 \\
\hline
\end{tabular}
\end{table*}

\begin{table*}[htbp]
\centering
\caption{Performance Metrics for Five-Feature Combination}
\label{tab:five-features}
\begin{tabular}{|c|l|cc|cc|cc|cc|cc|}
\hline
\multirow{2}{*}{No.} & \multirow{2}{*}{Features} & \multicolumn{2}{c|}{k-NN} & \multicolumn{2}{c|}{MLR} & \multicolumn{2}{c|}{DTC} & \multicolumn{2}{c|}{RF} & \multicolumn{2}{c|}{Average} \\
 &  & Acc \% & F1 & Acc \% & F1 & Acc \% & F1 & Acc \% & F1 & Acc \% & F1 \\
\hline
31 & RSSI+SNR+Frequency+Distance+Height & 66.48 & 66.76 & 60.33 & 59.19 & 68.04 & 68.05 & 68.05 & 67.99 & 65.73 & 65.50 \\
\hline
\end{tabular}
\end{table*}

%This may be because, in LoRaWAN networks, the carrier frequency is a fixed parameter that varies primarily by region. For instance, in Europe, the frequency bandwidth is set between 863 MHz to 870 MHz. \cite{1STBOOK} Given its limited variation within a specific deployment, the frequency may be leading to lower performance.

%On careful examination of the scatter plot in Figure 2, a general trend is observed where accuracy and F1 score tend to decrease within each group of feature combinations (e.g. 1 feature combination serial number 1-5, 2 feature combination 6-15 and so on). However, as the number of features in the combinations increases, an upward trend is observed in accuracy and F1 scores, with the plots gradually moving higher.

\begin{figure}
    \label{fig:b}
    \centering
    \includegraphics[width=0.7\linewidth]{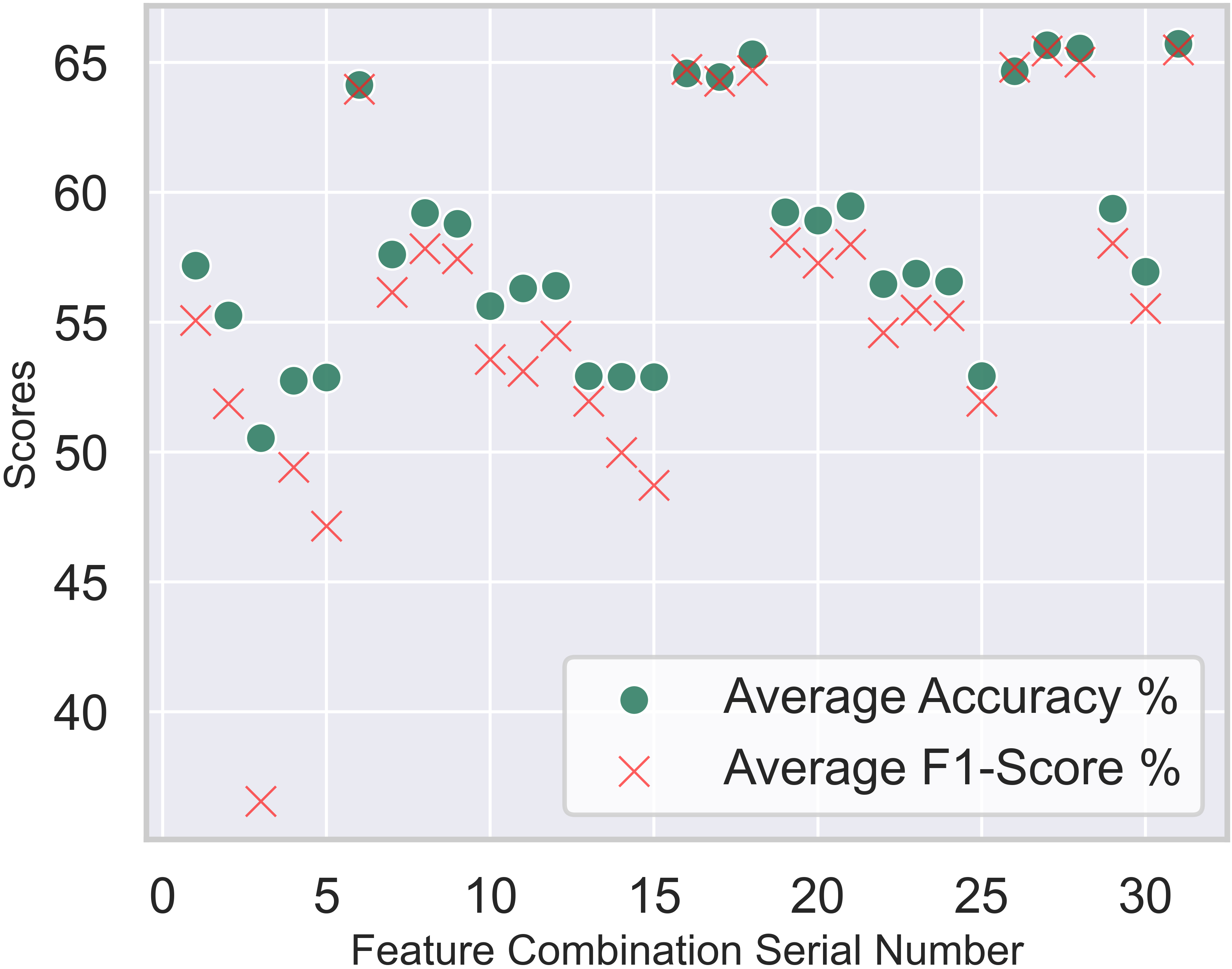}  
     \caption{\textbf{Visualization of metrics: Average Accuracy \% and F1 \%} }
    % if no solution works out then split the image in half and paste 2 image
\end{figure}

\begin{figure}[b!]
    
    \centering
    \includegraphics[width=0.7\linewidth]{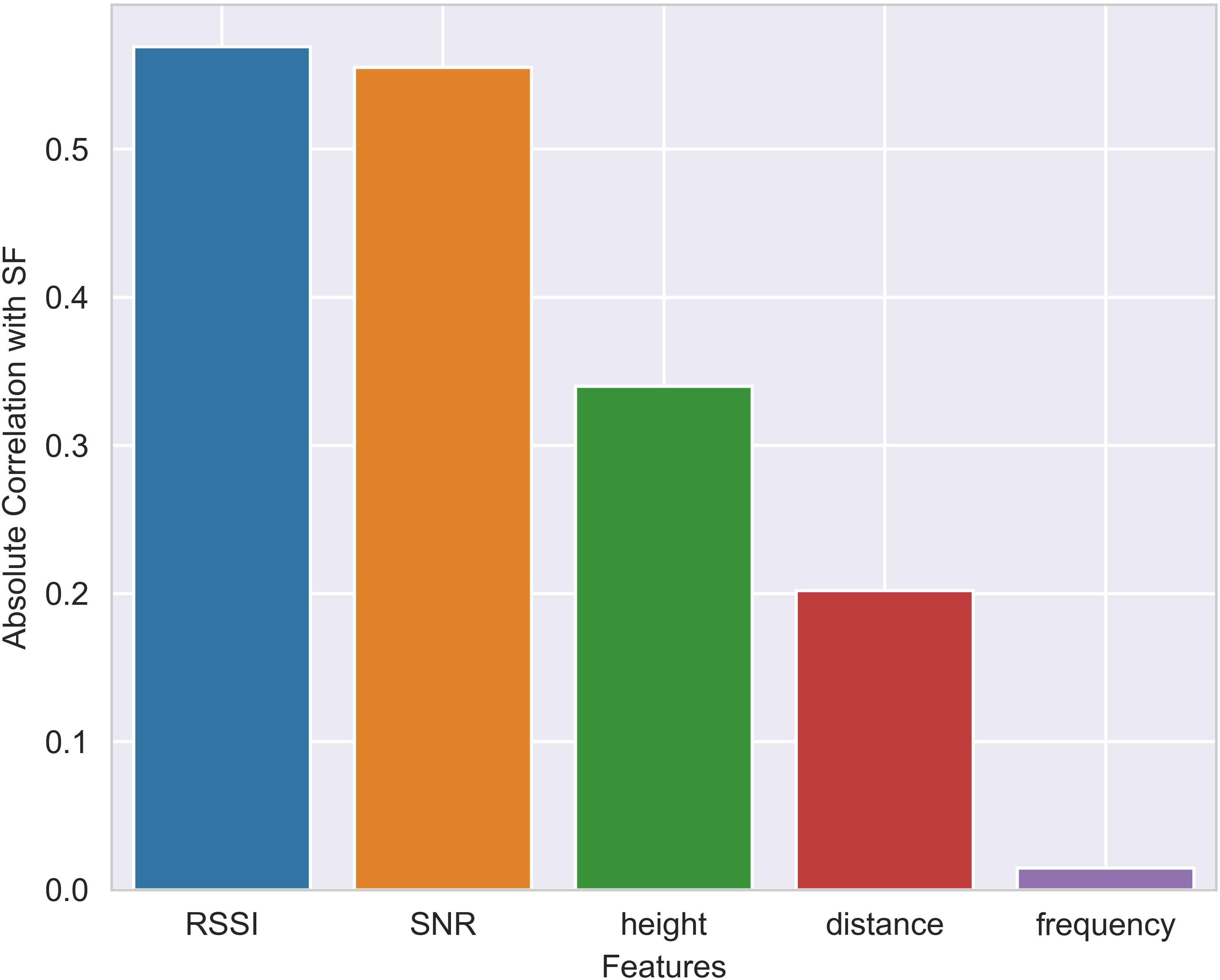}  
    \caption{\textbf{ Feature Ranking Based on Pearson Correlation} }
    \label{fig33}
\end{figure}

\subsection{Algorithm Performance}
DTC and RF consistently outperformed k-NN and MLR in both accuracy and F1 scores across all feature combinations. MLR consisted of the lowest metric among all four algorithms. However, The relative performance of feature combinations is consistent across all algorithms, suggesting feature combinations are not algorithm-dependent.

% Working here
\subsection{Verification of result using Filter Method}

 Pearson's correlation is a statistical method under the filter method, measures the linear relationship between features and the target variable \cite{PCC-Filter}. The result obtained from this method aligns with the findings we obtained in section \ref{feature analysis}. Specifically, RSSI has demonstrated the highest correlation and frequency the lowest, with the spreading factor (SF) as shown in Figure \ref{fig33}.

%plt.title('Feature Ranking Based on Correlation with SF')

%footer
% \fancypagestyle{plain}{    % Create a custom page style named "plain"
%     \fancyfoot[C]{\small\textit{Feature combination serial numbers are provided in Tables III-VII}}  % Centered footer text}
%\thispagestyle{plain}                          % Apply the "plain" style to the current page

%\clearpage%STARTS THE REST CONTENT FROM NEW PAGE

%\twocolumn[

%\small\textit{In the below tables, "Distance" refers to the distance between the GW and EDs, while "Height" denotes the ED antenna height. "Acc" stands for accuracy.}
%]

\subsection{Significance of Findings}

%IEEE24; rewritten below para
%REVISE {Significance of Findings}
\begin{comment}

    The findings of this paper have a significant impact on cost efficiency and battery life in LoRaWAN networks. The experimental setup for the collection of the dataset could simpler hardware design to skip the less important features such as frequency and focus only on collecting important features like RSSI, SNR, etc thus reducing the cost in data collection for model training.

The ML model trained only on combination of RSSI+SNR can reduce model training time and computation resource requirement.
Consequently, a model trained and deployed using the selected feature would require fewer computation resources to run, especially when deployed on end devices with limited processing capability \cite{lastcite}. After the model is trained on a reduced feature set and deployed in a live LoRaWAN environment, only the essential data needs to be collected for SF prediction, thereby reducing the load on LoRa devices and hence extending battery life.
\end{comment}

The results of this study have important implications for optimizing LoRaWAN network deployments and operations. The
%suggestion needed experimental setup or hardware setup IEEE24
experimental setup for the collection of the dataset can potentially be simpler in design rather than maintaining complex multi-sensor arrays to skip the less important features such as frequency and focus only on collecting important features like RSSI and SNR thus reducing the cost of data collection for model training.

The reduced feature set (RSSI + SNR) would require lower training time and computational requirements for ML model training.
Consequently, a model deployed using the selected feature would require lower memory and limited processing capability for inference, especially important for end devices with limited processing capability\cite{lastcite}. After deployment in a live LoRaWAN environment, only the essential new data needs to be collected for SF prediction, thereby reducing power consumption during data collection and processing, and contributing to extended battery life.

These advantages tackle key challenges in LoRaWAN networks requiring a long-term, battery-powered operation. This finding suggests, that focusing on these two features would provide cost-effective and energy-efficient implementations while maintaining good outcomes in SF prediction.

\section{Conclusion \& Future Work}

\label{conclude}

In this paper, we explored the influence of various feature
combinations in Spreading Factor prediction for LoRaWAN
networks using machine learning methods. Our results signify the importance of RSSI and SNR as key features, with their combination consistently outperforming other feature sets across four machine-learning algorithms. The result suggests that focusing on critical features can lead to efficient data collection and model training processes, further reducing computational requirements and extending battery life in IoT devices.

As a part of future work, we can further improve SF prediction leading to a reliable LoRaWAN network for the growing IoT ecosystem:
\begin{enumerate}
    \item \textbf{Exploring Deep Learning Techniques:} Neural networks can capture more complex relationships among the features and SF, which could potentially lead to higher accuracy and other improved metric scores.
\begin{comment}
\item \textbf{Expanding Feature Set:} In addition to the five features employed in this research, other features could be included to explore new insights. (e.g., Time On Air, Temperature, geographical data, etc.).
    
\end{comment}
    
    \item \textbf{Diverse dataset:} Experimenting with different rural or urban datasets could give more insights into this research.
\end{enumerate}

 \section*{Acknowledgement}
  This work is supported by the Science and Engineering Research Board, Department of Science and Technology, Government of India through Startup Research Grant, under Grant SRG/2023/002016.

\bibliographystyle{IEEEtran}
\bibliography{ref}

\end{document}